\documentclass[10pt,twocolumn,letterpaper]{article}

\usepackage{cvpr}
\usepackage{times}
\usepackage{epsfig}
\usepackage{graphicx}
\usepackage{amsmath}
\usepackage{amssymb}

\usepackage[english]{babel}
\usepackage[utf8]{inputenc}
\usepackage{algorithm}
\usepackage[noend]{algpseudocode}
\usepackage{nccmath}
\usepackage{multirow}
\usepackage{makecell}
\usepackage{boldline}

\usepackage[breaklinks=true,bookmarks=false]{hyperref}

\cvprfinalcopy 

\def\cvprPaperID{****} 

\begin{document}

\title{W-Net: A Deep Model for Fully Unsupervised Image Segmentation}

\author{Xide Xia\\
Boston University\\
{\tt\small xidexia@bu.edu}
\and
Brian Kulis\\
Boston University\\
{\tt\small bkulis@bu.edu}
}

\maketitle


\begin{abstract}
While significant attention has been recently focused on designing supervised deep semantic segmentation algorithms for vision tasks, there are many domains in which sufficient supervised pixel-level labels are difficult to obtain.  In this paper, we revisit the problem of purely unsupervised image segmentation and propose a novel deep architecture for this problem.  We borrow recent ideas from supervised semantic segmentation methods, in particular by concatenating two fully convolutional networks together into an autoencoder---one for encoding and one for decoding.  The encoding layer produces a k-way pixelwise prediction, and both the reconstruction error of the autoencoder as well as the normalized cut produced by the encoder are jointly minimized during training.  When combined with suitable postprocessing involving conditional random field smoothing and hierarchical segmentation, our resulting algorithm achieves impressive results on the benchmark Berkeley Segmentation Data Set, outperforming a number of competing methods.

\end{abstract}

\section{Introduction}

The image segmentation problem is a core vision problem with a longstanding history of research.  Historically, this problem has been studied in the unsupervised setting as a clustering problem: given an image, produce a pixelwise prediction that segments the image into coherent clusters corresponding to objects in the image.  In classical computer vision, there are a number of well-known techniques for this problem, including normalized cuts ~\cite{mnc,ncut}, Markov random field-based methods~\cite{zhang2001segmentation} , mean shift ~\cite{MS}, hierarchical methods ~\cite{arbelaez2011contour}, and many others.

\begin{figure}[h!]
\centering
\includegraphics[width=85mm]{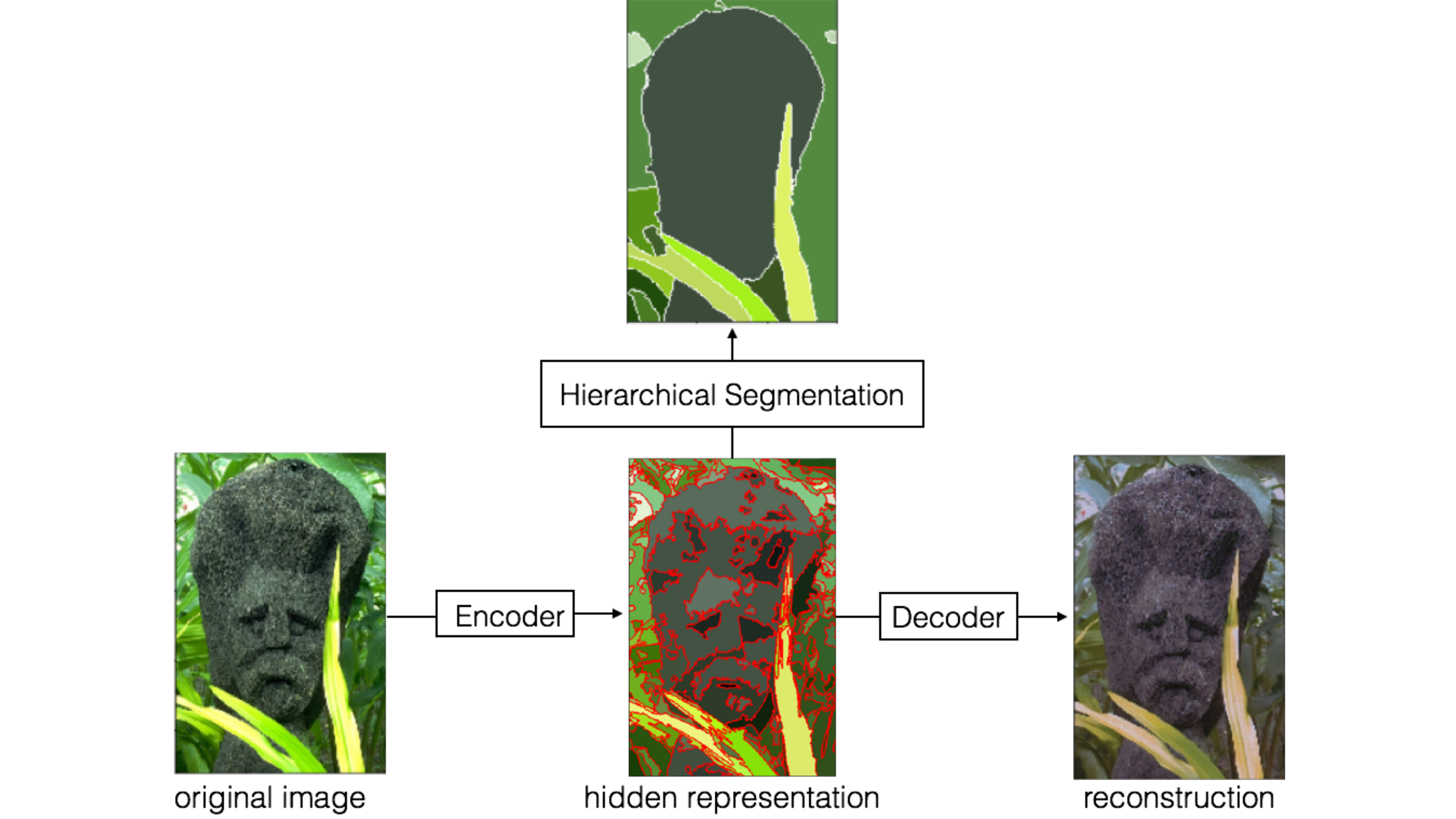}
\caption{\textbf{Overview of our approach.}  A fully convolutional network encoder produces a segmentation.  This segmentation is fed into a fully convolutional network decoder to produce a reconstruction, and training jointly minimizes the normalized cut of the encoded segmentation and the reconstruction of the image.  The encoded image is then post-processed to produce the final segmentation.}
\label{fig:autoencoder}
\end{figure}

Given the recent success of deep learning within the computer vision field, there has been a resurgence in interest in the image segmentation problem.  The vast majority of recent work in this area has been focused on the problem of \textit{semantic segmentation} ~\cite{badrinarayanan2015segnet,chaurasia2017linknet,paszke2016enet,unet,krahenbuhl2011efficient,zheng2015conditional}, a supervised variant of the image segmentation problem.  Typically, these methods are trained using models such as fully convolutional networks to produce a pixelwise prediction, and supervised training methods can then be employed to learn filters to produce segments on novel images.  One such popular recent approach is the U-Net architecture~\cite{unet}, a fully convolutional network that has been used to achieve impressive results in the biomedical image domain.  Unfortunately, existing semantic segmentation methods require a significant amount of pixelwise labeled training data, which can be difficult to collect on novel domains.

Given the importance of the segmentation problem in many domains, and due to the lack of supervised data for many problems, we revisit the problem of unsupervised image segmentation, utilizing recent ideas from semantic segmentation.  In particular, we design a new architecture which we call W-Net, and which ties two fully convolutional network (FCN) architectures (each similar to the U-Net architecture) together into a single autoencoder.  The first FCN encodes an input image, using fully convolutional layers, into a k-way soft segmentation.  The second FCN reverses this process, going from the segmentation layer back to a reconstructed image.  We jointly minimize both the reconstruction error of the autoencoder as well as a ``soft’’ normalized cut loss function on the encoding layer.  In order to achieve state-of-the-art results, we further appropriately postprocess this initial segmentation in two steps: we first apply a fully conneceted conditional random field (CRF) ~\cite{krahenbuhl2011efficient,chen2016deeplab} smoothing on the outputted segments, and we second apply the hierarchical merging method of~\cite{arbelaez2011contour} to obtain a final segmentation, showed as Figure \ref{fig:autoencoder}.

We test our method on the Berkeley Segmentation Data set benchmark.  We follow standard benchmarking practices and compute the segmentation covering (SC), probabilistic Rand index (PRI), and variation of information (VI) metrics of our segmentations as well as segmentation of several existing algorithms. We compare favorably with a number of classical and recent segmentation approaches, and even approach human-level performance in some cases---for example, our algorithm achieves 0.86 PRI versus 0.87 by humans, in the optimal image scale setting.  We further show several examples of segments produced by our algorithm as well as some competing methods.

The rest of this paper is organized as follows. We first review related works in Section 2. The architecture of our network is described in Section 3. Section 4  presents the detailed procedure of the post-processing method, and experimental results are demonstrated in Section 5. Section 4 discusses the conclusions that have been drawn.

\section{Related Work}
We briefly discuss related work on segmentation, convolutional networks, and autoencoders.

\subsection{Unsupervised Segmentation}

Most approaches to unsupervised image segmentation involve utilizing features such as color, brightness, or texture over local patches, and then make pixel-level clustering based on these features. Among these schemes, the three most widely-used methods include Felzenszwalb and Huttenlocher's graph-based method~\cite{Felz-Hutt}, Shi and Malik's Normalized Cuts ~\cite{mnc,ncut}, and Comaniciu and Meer's Mean Shift~\cite{MS}. Arbelaez et al.~\cite{arbelaez2009contours,arbelaez2011contour} proposed a method based on edge detection that has been shown to outperform the classical methods. More recently, \cite{pont2017multiscale} proposed a unified approach for bottom-up multi-scale hierarchical image segmentation. In this paper, we adopt the hierarchical grouping algorithm described in ~\cite{arbelaez2011contour} for postprocessing after we get an initial segmentation prediction from W-Net encoder.

\subsection{Deep CNNs in Semantic Segmentation}

Deep neural networks have emerged as a key component in many visual recognition problems, including supervised learning for semantic image segmentation. \cite{mostajabi2015feedforward,farabet2013learning,dai2015convolutional,hariharan2014simultaneous,hariharan2015hypercolumns} all make pixel-wise annotations for segmentation based on supervised classification using deep networks.
 
Fully convolutional networks (FCNs)~\cite{fcn} have emerged as one of the most effective models for the semantic segmentation problem. In a FCN, fully connected layers of standard convolutional neural networks (CNNs) are transformed as convolution layers with kernels that cover the entire input region. By utilizing fully connected layers, the network can take an input of arbitrary size and produce a correspondingly-sized output map; for example, one can produce a pixelwise prediction for images of arbitrary size. Recently a number of variants of FCN have been proposed and studied and that perform semantic segmentation~\cite{noh2015learning,badrinarayanan2015segnet,chaurasia2017linknet,paszke2016enet,unet,krahenbuhl2011efficient,zheng2015conditional}. In \cite{krahenbuhl2011efficient}, a conditional random field (CRF) is applied to the output map to fine-tune the segmentation. \cite{zheng2015conditional} formulates a mean-field approximate inference for the CRF as a Recurrent Neural Network (CRF-RNN), and then jointly optimize both the CRF energy as well as the supervised loss. \cite{unet} presents a U-shaped architecture consisting of a contracting path to capture context and a symmetric expanding path that enables precise localization. In this paper, we modify and extend the architecture described in~\cite{unet} to a W-shaped network such that it reconstructs the original input images and also predicts a segmentation map without any labeling information.

\subsection{Encoder-decoders}
 
Encoder-decoders are one of the most widely known and used methods in unsupervised feature learning ~\cite{AE1,AE2}. The encoder $\mathbf{Enc}$ maps the input (\eg an image patch) to a compact feature representation, and then the decoder $\mathbf{Dec}$ reproduces the input from its lower-dimensional representation.  In this paper, we design an encoder such that the input is mapped to a dense pixelwise segmentation layer with same spatial size rather than a low-dimensional space.  The decoder then performs a reconstruction from the dense prediction layer.

\begin{figure*}
\begin{center}
\includegraphics[width=185mm]{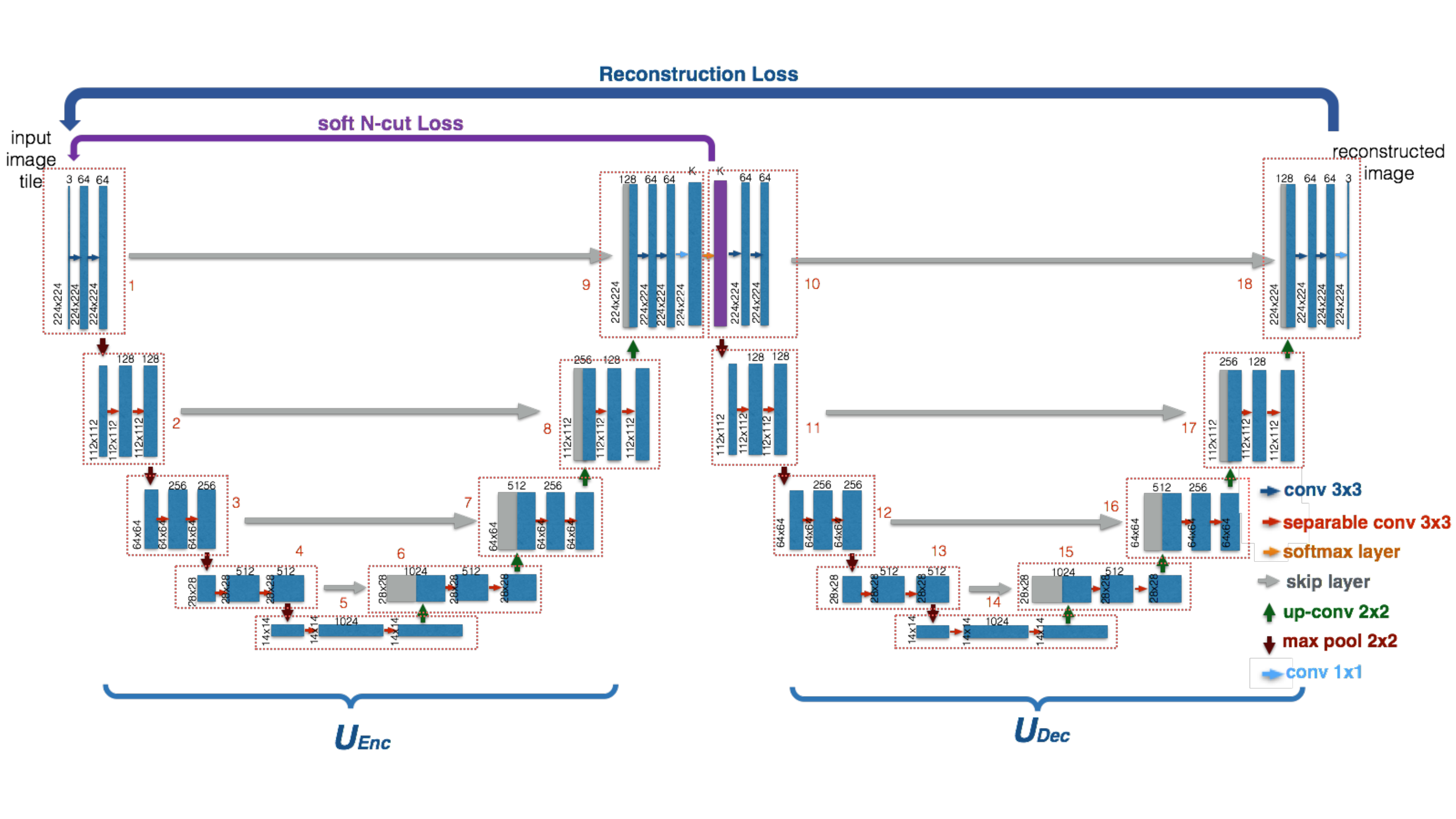}
\end{center}
   \caption{\textbf{W-Net architecture.} The W-Net architecture is consist of an $U_{Enc}$ (left side) and a corresponding $U_{Dec}$ (right side). It has 46 convolutional layers which are structured into 18 modules marked with the red rectangles. Each module consists of two 3 $\times$ 3 convolutional layers. The first nine modules form the dense prediction base of the network and the second 9 correspond to the reconstruction decoder.}
\label{fig:wnet}
\end{figure*}

\section{Network Architecture}
The network architecture is illustrated in Figure ~\ref{fig:wnet}.  It is divided into an $U_{Enc}$ (left side) and a corresponding $U_{Dec}$ (right side); in particular, we modify and extend the typical U-shaped architecture of a U-Net network described in~\cite{unet} to a W-shaped architecture such that it reconstructs original input images as well as predicts the segmentation maps without any labeling information. The W-Net architecture has 46 convolutional layers which are structured into 18 modules marked with the red rectangles. Each module consists of two 3 $\times$ 3 convolutional layers, each followed by a ReLU ~\cite{relu}  non-linearity and batch normalization~\cite{bn}.  The first nine modules form the dense prediction base of the network and the second 9 correspond to the reconstruction decoder.

The $U_{Enc}$ consists of a contracting path (the first half) to capture context and a corresponding expansive path (the second half) that enables precise localization, as in the original U-Net architecture. The contracting path starts with an initial module which performs convolution on input images.  In the figure, the output sizes are reported for an example input image resolution of 224 $\times$ 224. Modules are connected via 2 $\times$ 2 max-pooling layers, and we double the number of feature channels at each downsampling step. In the expansive path, modules are connected via transposed 2D convolution layers.  We halve the number of feature channels at each upsampling step. As in the U-Net model, the input of each module in the contracting path is also bypassed to the output of its corresponding module in the expansive path to recover lost spatial information due to downsampling. The final convolutional layer of the $U_{Enc}$ is a 1 $\times$ 1 convolution followed by a softmax layer. The 1x1 convolution maps each 64-component feature vector to the desired number of classes $K$, and then the softmax layer rescales them so that the elements of the $K$-dimensional output lie in the range (0,1) and sum to 1. The architecture of the $U_{Dec}$ is similar to the $U_{Enc}$ except it reads the output of the $U_{Enc}$ which has the size of 224 $\times$ 224 $\times$ $K$. The final convolutional layer of the $U_{Dec}$ is a 1 $\times$ 1 convolution to map 64-component feature vector back to a reconstruction of original input.

One important modification in our architecture is that all of the modules use the depthwise separable convolution layers introduced in~\cite{xception} except modules 1, 9, 10, and 18. A depthwise separable convolution operation consists of a depthwise convolution and a pointwise convolution. The idea behind such an operation is to examine spatial correlations and cross-channel correlations independently---a depthwise convolution performs spatial convolutions independently over each channel and then a pointwise convolution projects the feature channels by the depthwise convolution onto a new channel space. As a consequence, the network gains performance more efficiently with the same number of parameters. In Figure~\ref{fig:wnet}, blue arrows represent convolution layers and red arrows indicate depth-wise separable convolutions. The network does not include any fully connected layers which allow it to learn arbitrarily large images and make a segmentation prediction of the corresponding size.

\subsection{Soft Normalized Cut Loss}

The output of the $U_{Enc}$ is a normalized 224 $\times$ 224 $\times$ $K$ dense prediction. By taking the argmax, we can obtain a $K$-class prediction for each pixel. In this paper, we compute the normalized cut ($Ncut$)~\cite{ncut} as a global criterion for the segmentation:
\begin{equation}\label{nc_loss}
\begin{gathered}
	Ncut_{K}(V) 
    = \sum_{k=1}^{K} \frac{cut(A_{k},V-A_{k})}{assoc(A_k,V)} \\
    = \sum_{k=1}^{K} \frac{\sum_{u\in A_{k},v\in V-A_{k}} w(u,v)}{\sum_{u\in  A_{k},t\in V} w(u,t)}, \\
\end{gathered}
\end{equation}
where $A_k$ is set of pixels in segment $k$, $V$ is the set of all pixels, and $w$ measures the weight between two pixels. 

However, since the argmax function is non-differentiable, it is impossible to calculate the corresponding gradient during backpropagation. Instead, we define a \textit{soft} version of the $Ncut$ loss which is differentiable so that we can update gradients during backpropagation:
\begin{ceqn}
\begin{align}\label{soft_nc_loss}
\begin{gathered}
	J_{soft-Ncut}(V,K) 
    = \sum_{k=1}^{K} \frac{cut(A_{k},V-A_{k})}{assoc(A_k,V)} \\
    = K -\sum_{k=1}^{K} \frac{assoc(A_{k},A_{k})}{assoc(A_k,V)} \\
    = K - \sum_{k=1}^{K} \frac{\sum_{u\in V,v\in V} w(u,v)p(u= A_{k})p(v= A_{k})} {\sum_{u\in  A_{k},t\in V} w(u,t)p(u= A_{k})} \\
     = K - \sum_{k=1}^{K} \frac{\sum_{u\in V}p(u= A_{k}) \sum_{u\in V}w(u,v)p(v= A_{k})} {\sum_{u\in V}p(u= A_{k}) \sum_{t\in V}w(u,t)},
\end{gathered}
\end{align}
\end{ceqn}
where $p(u= A_{k})$ measures the probability of node $u$ belonging to class $A_{k}$, and which is directly computed by the encoder. By training $U_{Enc}$ to minimize the $J_{soft-Ncut}$ loss we can simultaneously minimize the total normalized disassociation between the groups and maximize the total normalized association within the groups.

\subsection{Reconstruction Loss}
As in the classical encoder-decoder architecture, we also train the W-Net to minimize the reconstruction loss to enforce that the encoded representations contain as much information of the original inputs as possible. In this paper, by minimizing the reconstruction loss, we can make the segmentation prediction align better with the input images.  The reconstruction loss is given by
\begin{equation}\label{rec_loss}
	J_{reconstr} =\left \|  \mathbf{X} - \mathbf{U_{Dec}}(\mathbf{U_{Enc}}(\mathbf{X};W_{Enc});W_{Dec}) \right \|_{2}^{2},
\end{equation}
where $W_{Enc}$ denotes the parameters of the encoder, $W_{Dec}$ denotes the parameters of the decoder, and $\mathbf{X}$ is the input image.
We train W-Net to minimize the $J_{reconstr}$ between the reconstructed images and original inputs. We simultaneously train $U_{Enc}$ to minimize $J_{soft-Ncut}$ in order to maximize the association within segments and minimize the disassociation between the segments in the encoding layer. The procedure is formally presented in Algorithm 1. By iteratively applying $J_{reconstr}$ and $J_{soft-Ncut}$, the network balances the trade-off between the accuracy of reconstruction and the consistency in the encoded representation layer. 

\begin{algorithm}
\caption{Minibatch stochastic gradient descent training of W-Net.}\label{alg:euclid}
\begin{algorithmic}[1]
\Procedure{W-Net}{$\mathbf{X}$;$U_{Enc}, U_{Dec}$}  
\For{ number of training iterations}
        \State Sample a minibatch of new input images $x$
        \State Update $U_{Enc}$ by minimizing $J_{soft-Ncut}$   \\
        \Comment{Only update $U_{Enc}$}
        \State Update whole W-Net by minimizing $J_{reconstr}$     \\
        \Comment{Update both $U_{Enc}$ and $U_{Dec}$}
\EndFor
\State \textbf{return} $U_{Enc}$
\EndProcedure
\label{algo:wnet22}
\end{algorithmic}
\end{algorithm}

\section{Postprocessing}
After obtaining an initial segmentation from the encoder, we perform two postprocessing steps in order to obtain our final result.  Below we describe these steps, namely CRF smoothing and hierarchical merging.

\subsection{Fully-Connected Conditional Random Fields for Accurate Edge Recovery}

While deep CNNs with max-pooling layers have proven their success in capturing high-level feature information of inputs, the increased invariance and large receptive fields can cause reduction of localization accuracy. A lack of smoothness constraints can result in the problem of poor object delineation, especially in pixel-level labeling tasks.

To address this issue, while the soft normalized cut loss and skip layers in the W-Net can help to improve the localization of object boundaries, we find that it improves segmentations with fine-grained boundaries by combining the responses at the final $U_{Enc}$ layer with a fully connected Conditional Random Field (CRF) model~\cite{chen2016deeplab}. 
The fully connected CRF model employs the energy function
\begin{equation}\label{rec_loss}
	E(\mathbf{X}) =  \sum_{u}\Phi  (u) +  \sum_{u,v}\Psi (u,v)
\end{equation}
where $u,v$ are pixels on input data $\mathbf{X}$. The unary potential $\Phi  (u) = - \log p(u)$, where $ p(u)$ is the label annotation probability computed by the softmax layer in $U_{Enc}$. The pairwise potential $\Psi (u,v)$ measures the weighted penalties when two pixels are assigned different labels by using two Gaussian kernels in different feature spaces.

\begin{figure}[h!]
\centering
\includegraphics[width=85mm]{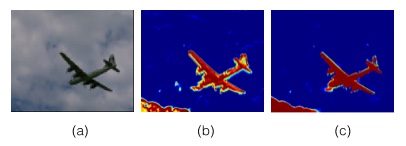}
\caption{ \textbf{Belief map (output of softmax function) before and after a fully connected CRF model.} (a) Original image, (b) the responses at the final $U_{Enc}$ layer, (c) the output of the fully connected CRF.}
\label{fig:fc_crf}
\end{figure}

Figure~\ref{fig:fc_crf} presents an example of the prediction before and after the fully connected CRF model. The output of the softmax layer in the fully convolutional encoder $U_{Enc}$ predicts the rough position of objects in inputs with coarse boundaries. After the fully connected CRF model, the boundaries are sharper and small spurious regions have been smoothed out or removed.

\subsection{Hierarchical Segmentation}
\begin{figure}[h!]
\centering
\includegraphics[width=82mm]{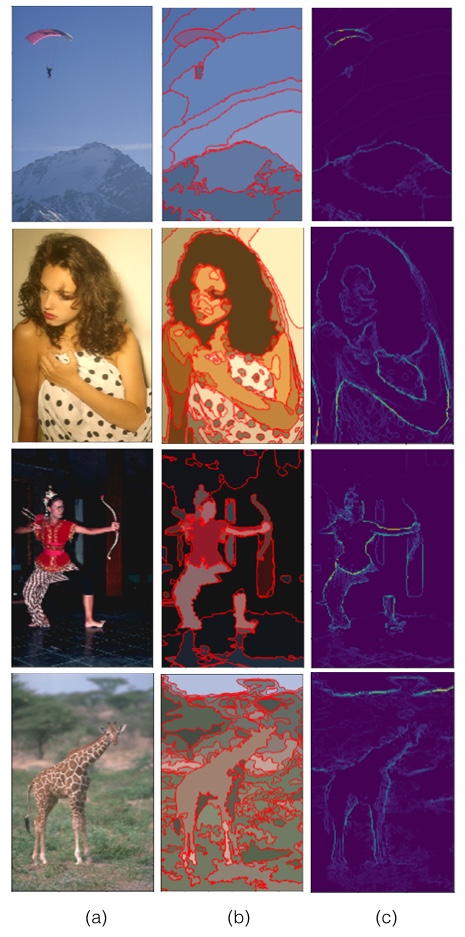}
\caption{ \textbf{The initial partitions for the hierarchical merging and the corresponding weighted boundary maps.} (a) Original inputs. (b) The output (argmax) of the fully connected CRF with boundaries showed in red lines. (c) The corresponding weighted boundary maps of the red lines in (b).}
\label{fig:seg2crf2edge}
\end{figure}

After taking the argmax on the output of the fully connected CRF, we still typically obtain an over-segmented partition of the input image.  Our final step is to merge segments appropriately to form the final image segments.  Figure~\ref{fig:seg2crf2edge} shows examples of such initial regions with boundaries in red lines for original input images in (a). In this section, we discuss an efficient hierarchical segmentation that first converts the over-segmented partitions into weighted boundary maps and then merges the most similar regions iteratively.

We measure the ``importance" of each pixel on the initial over-segmented partition boundaries by computing a weighted combination of multi-scale local cues and global boundary measurements based on spectral clustering~\cite{arbelaez2009contours,arbelaez2011contour}:
\begin{equation}\label{gPb}
	gPb(x,y,\theta) =  \sum_{s}\sum_{i}\beta_{i,s} G_{i,\sigma(s)}(x,y,\theta) \\
    + \gamma sPb(x,y,\theta),
\end{equation}
where $s$ indexes scales, $i$ indexes feature channels including brightness, color, and texture, and $G_{i,\sigma(s)}(x,y,\theta)$ measures the dissimilarity  between two halves of a disc of radius $\sigma(s)$ center at $(x,y)$ in channel $i$ at angle $\theta$. The $mPb$ signal measures all the edges in the image and the $sPb$ signal captures the most salient curves in the image. Figure~\ref{fig:seg2crf2edge} (c) shows the corresponding weighted boundary maps of the initial boundaries produced by WNet-CRF in Figure~\ref{fig:seg2crf2edge} (b).
       
We then build hierarchical segmentation from this weighted boundaries by using the countour2ucm stage described in~\cite{arbelaez2009contours,arbelaez2011contour}. This algorithm constructs a hierarchy of segments from contour detections. It has two steps: an Oriented Watershed Transform (OWT) to build an initial over-segmented region and an Ultrametric Contour Map (UCM), which is a greedy graph-based region merging algorithm. 

\begin{algorithm}
\caption{Post processing}\label{alg:euclid}
\begin{algorithmic}[1]
\Procedure{Postprocessing}{$\mathbf{x}$;$U_{Enc}, CRF, Pb$}
\State   ${x}' = U_{Enc}(x)$   \\
\Comment{Get the hidden representation of $x$}
\State $ {x}^{''}=\textsc{CRF}({x}')$    \\
\Comment{fine-grained boundaries with a fully CRF}
\State $ x^{'''} = Pb({x}^{''})$    \\
\Comment{compute the probability of boundary only on the edge detected in ${x}^{''}$}
\State $ \mathbf{\mathit{S}} = countour2ucm({x}^{'''})$    \\
\Comment{hierarchical segmentation}
\State \textbf{return} $\mathbf{\mathit{S}}$
\EndProcedure
\end{algorithmic}
\end{algorithm}

\section{Experiments}

We train our proposed W-Net on the PASCAL VOC2012 dataset~\cite{voc2012} and then evaluate the trained network using the Berkeley Segmentation Database (BSDS300 and BSDS500). The PASCAL VOC2012 dataset is a large visual object classes challenge which contains 11,530 images and 6,929 segmentations. BSDS300 and BSDS500 have 300 and 500 images, respectively. For each image, the BSDS dataset provides human-annotated segmentation as ground truth.  Since our proposed method is designed for unsupervised image segmentation, we do not use any ground truth labels in the training phase; we use the ground truth only to evaluate quality of our segmentations.

We resize the input images to 224 $\times$ 224 during training, and the architecture of the trained network is shown in Figure~\ref{fig:wnet}. We train the networks from scratch using mini-batches of 10 images, with an initial learning rate of 0.003. The learning rate is divided by ten after every 1,000 iterations. The training is stopped after 50,000 iterations. Dropout of 0.65 was added to prevent overfitting during training. We construct the weight matrix $W$ for $J_{soft-Ncut}$ as:
\begin{ceqn}
\scriptsize
\begin{align}\label{NCweights}
\begin{gathered}
w_{ij} = e^{\frac{-\left \| F(i)-F(j) \right \|_{2}^{2}}{\sigma_I^2} } 
    \ast \left\{\begin{matrix}
e^{\frac{-\left \| X(i)-X(j) \right \|_{2}^{2}}{\sigma_X^2} }  & \mbox{ if } \left \| X(i)-X(j) \right \|_{2} <r\\ 
0 & \mbox{otherwise},
\end{matrix}\right. 
\end{gathered}
\end{align}
\end{ceqn}

where $X(i)$ and $F(i)$ are the spatial location and pixel value of node $i$, respectively. $\sigma_I = 10$, $\sigma_X = 4$, and $r = 5$.

\begin{figure}[h!]
\centering
\includegraphics[width=90mm]{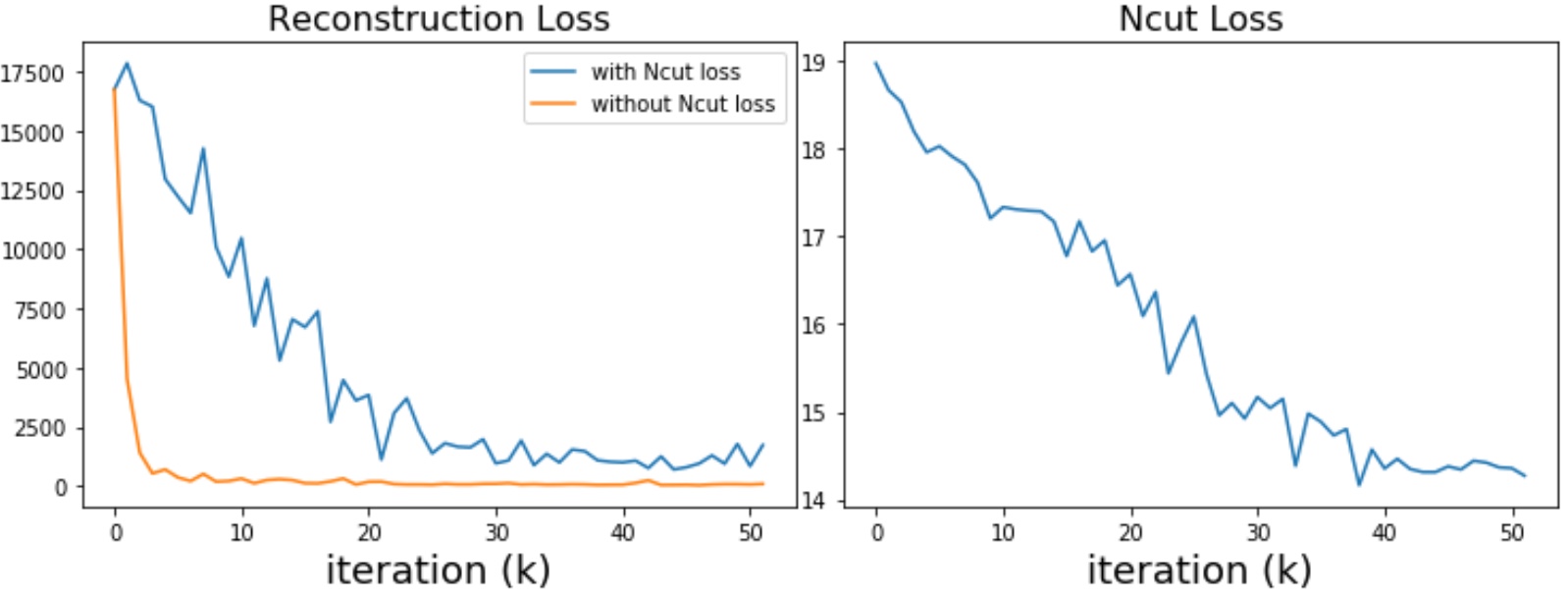}
\caption{ \textbf{ $J_{reconstr}$ and $J_{soft-Ncut}$  losses in the training phase. } \textbf{Left:} Reconstruction losses during training (red: training without $J_{soft-Ncut}$ , blue: training with $J_{soft-Ncut}$ ). \textbf{Right:} Soft-Ncuts loss during training.  }
\label{fig:loss_log}
\end{figure}

The plots of $J_{reconstr}$  and $J_{soft-Ncut}$ losses during training are shown in Figure~\ref{fig:loss_log}. We examine the $J_{reconstr}$ loss with and without considering $J_{soft-Ncut}$ during training. From Figure~\ref{fig:loss_log}, we can see that $J_{reconstr}$ converges faster when the $J_{soft-Ncut}$ is not considered. When we add the $J_{soft-Ncut}$ during training, the $J_{reconstr}$ decreases slowly and less stably. At convergence of $J_{reconstr}$, the blue line (training with $J_{soft-Ncut}$) is still higher than the red one (training without); this is because the hidden representation space is forced to be more consistent with a good segmentation of the image when the $J_{soft-Ncut}$ loss is introduced, so its ability to reconstruct the original images is weakened. Finally, both $J_{reconstr}$ and $J_{soft-Ncut}$  converge which means our approach balances trade-offs between minimizing the reconstruction loss in the last layer and maximizing the total association within the groups in the hidden layer. 

\begin{figure}[h!]
\centering
\includegraphics[width=85mm]{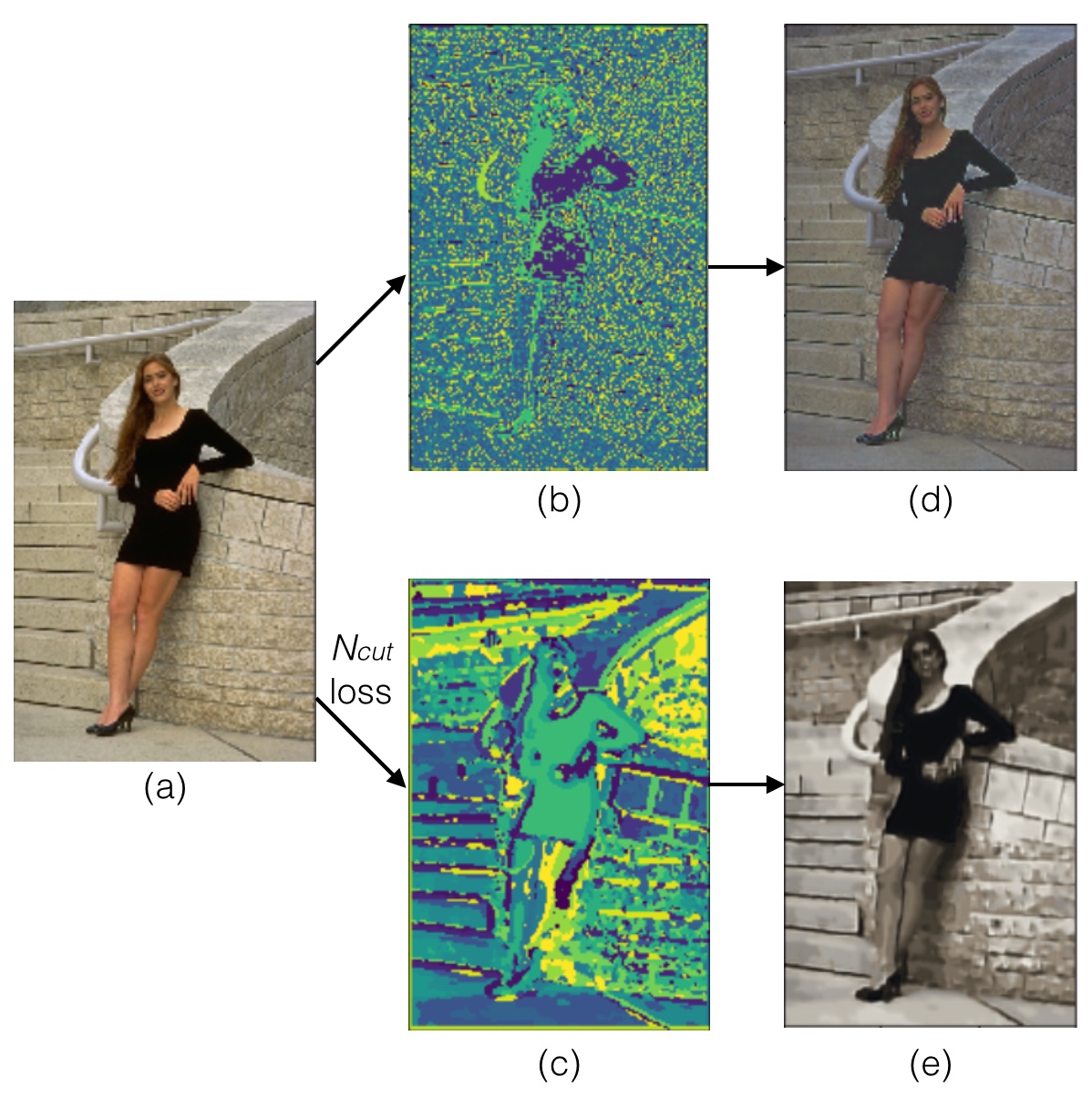}
   \caption{\textbf{A comparison with and without considering $J_{soft-Ncut}$ loss during back-propagation.} (a) Original image. (b) Visualization of the output of the final softmax layer in $U_{Enc}$ without adding $J_{soft-Ncut}$ loss during back-propagation. (c) Visualization of the output in the final softmax layer in $U_{Enc}$ when adding $J_{soft-Ncut}$ loss during back-propagation. (d) The corresponding reconstructed image of (b).  (e) The corresponding reconstructed image of (c).}
\label{fig:w_wout_NC}
\end{figure}

\begin{figure*}
\begin{center}
\includegraphics[width=\textwidth]{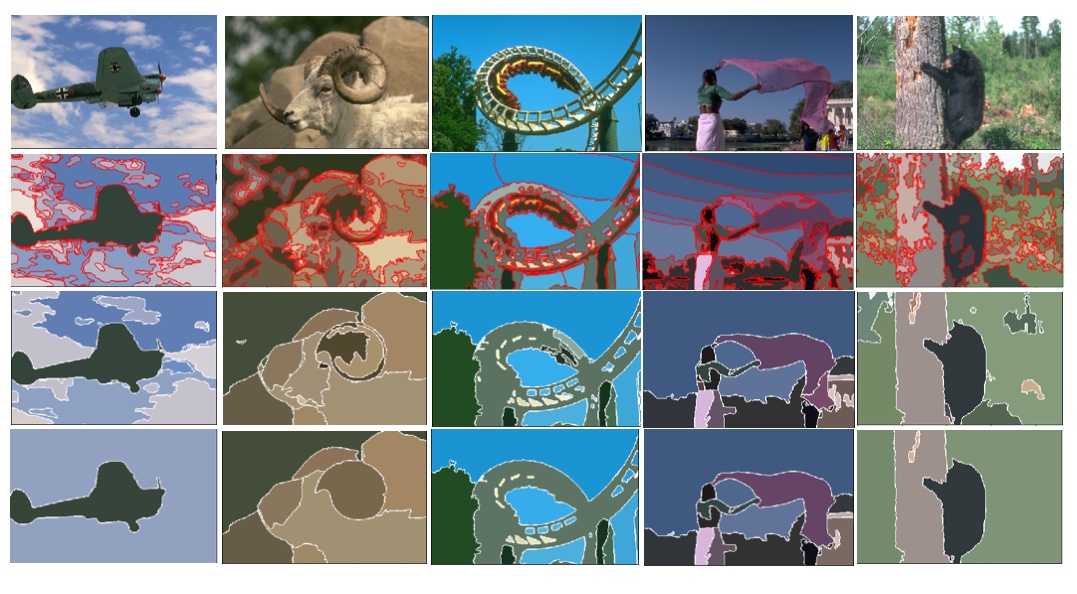}
\end{center}
   \caption{\textbf{Results of hierarchical segmentation using the output of WNet with CRF smoothing as initial boundaries, on the BSDS500.} From top to bottom: Original image, the initial over-segmented partitions showed in red lines obtained by the fully connected CRF, segmentations obtained by thresholding at the optimal dataset scale (ODS) and optimal image scale (OIS).}
\label{fig:no_ucm_result}
\end{figure*}

\begin{figure*}
\begin{center}
\includegraphics[width=\textwidth]{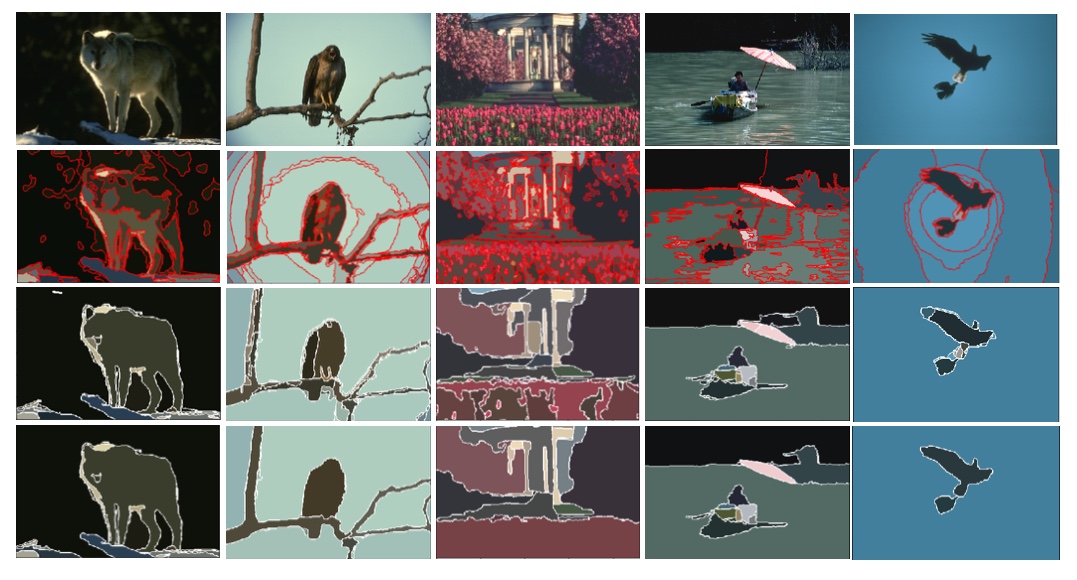}
\end{center}
   \caption{\textbf{Results of hierarchical segmentation using the combination of the output of WNet with CRF smoothing and UCM as the initial boundaries, on the BSDS500.} From top to bottom: Original image, the initial over-segmented partitions showed in red lines, segmentations obtained by thresholding at the optimal dataset scale (ODS) and optimal image scale (OIS).}
\label{fig:add_ucm_result}
\end{figure*}

Figure~\ref{fig:w_wout_NC} illustrates the comparison with and without considering $J_{soft-Ncut}$ loss during back-propagation. In order to make a better visualization of the output on the softmax layer in $U_{Enc}$, we take the argmax function on the prediction and use different colors to visualize different pixel-wise labels. We can see that the pixel-wise prediction is smoothed when we consider the $J_{soft-Ncut}$ during back-propagation. When we remove the $J_{soft-Ncut}$ loss from the W-Net, the model become a regular fully convolutional encoder-decoder which makes a high-quality reconstruction; however, the output of softmax layer is more noisy and discrete. On the other hand, by adding the $J_{soft-Ncut}$  loss, we get a more consistent hidden representation shown in (d), although the reconstruction is not as good as the one in a classical encoder-decoder architecture. From this comparison, we can see the trade-off between the consistency in the hidden representation and the quality of reconstruction, and it justifies our use of a soft normalized cut loss during training.

\subsection{Segmentation Benchmarks}

\begin{table}[]
\label{table:bsds300}
\resizebox{\columnwidth}{!}{%
\begin{tabular}{c|c|c|c|c|c|c}
\Xhline{1.0pt}
\multicolumn{1}{c|}{\multirow{2}{*}{Method}} & \multicolumn{2}{c|}{SC}                                            & \multicolumn{2}{c|}{PRI}                                                & \multicolumn{2}{c}{VI}                                                 \\ \cline{2-7} 
\multicolumn{1}{c|}{}                        & \multicolumn{1}{c|}{ODS}           & \multicolumn{1}{c|}{OIS}           & \multicolumn{1}{c|}{ODS}           & \multicolumn{1}{c|}{OIS}           & \multicolumn{1}{c|}{ODS}           & \multicolumn{1}{c}{OIS}           \\ \Xhline{1.0pt}
Quad Tree               & 0.33          & 0.39                & 0.71          & 0.75          & 2.34          & 2.22          \\ \hline
Chan Vese~\cite{chanvese}      & 0.49          & -                   & 0.75          & -             & 2.54          & -             \\ \hline
NCuts~\cite{mnc}           & 0.44          & 0.53                & 0.75          & 0.79          & 2.18          & 1.84          \\ \hline
SWA~\cite{SWA}            & 0.47          & 0.55                 & 0.75          & 0.80          & 2.06          & 1.75          \\ \hline
Canny-owt-ucm~\cite{arbelaez2011contour}  & 0.48          & 0.56              & 0.77          & 0.82          & 2.11          & 1.81          \\ \hline
Felz-Hutt~\cite{Felz-Hutt}       & 0.51          & 0.58               & 0.77          & 0.82          & 2.15          & 1.79          \\ \hline
Mean Shift~\cite{MS}     & 0.54          & 0.58              & 0.78          & 0.80          & 1.83          & 1.63          \\ \hline
Taylor~\cite{taylor2013towards}     & 0.56          & 0.62               & 0.79          & 0.84          & 1.74          & 1.63          \\ \hline
\textbf{W-Net (ours)}          & \textbf{0.58} & \textbf{0.62}  & \textbf{0.81} & \textbf{0.84} & \textbf{1.71} & \textbf{1.53} \\ \hline
gPb-owt-ucm~\cite{arbelaez2011contour}     & 0.59          & 0.65              & 0.81          & 0.85          & 1.65          & 1.47          \\ \hline
\textbf{W-Net+ucm (ours)}          & \textbf{0.60} & \textbf{0.65}  & \textbf{0.82} & \textbf{0.86} & \textbf{1.63} & \textbf{1.45} \\ \hline
Human                   & 0.73          & 0.73                 & 0.87          & 0.87          & 1.16          & 1.16          \\ \Xhline{1.0pt}
\end{tabular}%
}
\caption{\textbf{Results on BSDS300}. The values are reproduced from the tables in ~\cite{taylor2013towards}.}
\end{table}

\begin{table}[]
\label{table:bsds500}
\resizebox{\columnwidth}{!}{%
\begin{tabular}{c|c|c|c|c|c|c}
\Xhline{1.0pt}
\multicolumn{1}{c|}{\multirow{2}{*}{Method}} & \multicolumn{2}{c|}{SC}                                            & \multicolumn{2}{c|}{PRI}                                                & \multicolumn{2}{c}{VI}                                                 \\ \cline{2-7} 
\multicolumn{1}{c|}{}                        & \multicolumn{1}{c|}{ODS}           & \multicolumn{1}{c|}{OIS}           & \multicolumn{1}{c|}{ODS}           & \multicolumn{1}{c|}{OIS}           & \multicolumn{1}{c|}{ODS}           & \multicolumn{1}{c}{OIS}           \\ \Xhline{1.0pt}
NCuts~\cite{mnc}          & 0.45          & 0.53              & 0.78          & 0.80          & 2.23          & 1.89          \\ \hline
Canny-owt-ucm~\cite{arbelaez2011contour}   & 0.49          & 0.55              & 0.79          & 0.83          & 2.19          & 1.89          \\ \hline
Felz-Hutt~\cite{Felz-Hutt}      & 0.52          & 0.57              & 0.80          & 0.82          & 2.21          & 1.87          \\ \hline
Mean Shift~\cite{MS}    & 0.54          & 0.58             & 0.79          & 0.81          & 1.85          & 1.64          \\ \hline
Taylor~\cite{taylor2013towards}     & 0.56          & 0.62           & 0.81          & 0.85          & 1.78          & 1.56          \\ \hline
\textbf{W-Net (ours)}          & \textbf{0.57} & \textbf{0.62}  & \textbf{0.81} & \textbf{0.84} & \textbf{1.76} & \textbf{1.60} \\ \hline
gPb-owt-ucm~\cite{arbelaez2011contour}    & 0.59          & 0.65            & 0.83          & 0.86          & 1.69          & 1.48          \\ \hline
DC-Seg-full~\cite{donoser2014discrete}   & 0.59          & 0.64          & 0.82          & 0.85          & 1.68          & 1.54          \\ \hline
\textbf{W-Net+ucm (ours)}          & \textbf{0.59} & \textbf{0.64}  & \textbf{0.82} & \textbf{0.85} & \textbf{1.67} & \textbf{1.47} \\ \hline
Human                   & 0.72          & 0.72                 & 0.88          & 0.88          & 1.17          & 1.17          \\ \Xhline{1.0pt}
\end{tabular}%
}
\caption{\textbf{Results on BSDS500}. The values are reproduced from the tables in ~\cite{taylor2013towards} and ~\cite{donoser2014discrete}.}
\end{table}

To compare the performance of W-Net with existing unsupervised image segmentation methods, we compare with the following: DC-Seg-full~\cite{donoser2014discrete}, gPb-owt-ucm ~\cite{arbelaez2011contour}, Taylor~\cite{taylor2013towards}, Felzenszwalb and Huttenlocher (Felz-Hutt)~\cite{Felz-Hutt}, Mean Shift~\cite{MS}, Canny-owt-ucm ~\cite{arbelaez2011contour}, SWA~\cite{SWA}, Chan Vese~\cite{chanvese},  Multiscale Normalized Cuts (NCuts)~\cite{mnc}, and Quad-Tree. As has become standard, we evaluate the performance on three different metrics: Variation of Information (VI), Probabilistic Rand Index (PRI), and Segmentation Covering (SC).  For SC and PRI, higher scores are better; for VI, a lower score is better.  We also report human performance on this data set. For a set of hierarchical segmentations {$\mathbf{\mathit{S_i}}$} corresponding to different scales, we report the result at Optimal Dataset Scale (ODS) and Optimal Image Scale (OIS).

Table 1 and Table 2 summarize the performance of the proposed method noted as $\mathbf{W}$-$\mathbf{Net}$ on BSDS300 and BSDS500 respectively. Since the $U_{Enc}$ encoder followed by a fully connected CRF provides an initial boundaries detection, we compute the multi-scale local cues only on the detected edges instead of the whole input image.  As can be seen, our proposed approach has competitive performance compared to the high computation demanding gPb-owt-ucm method. We further consider combining the boundaries produced by our W-Net model after CRF smoothing with the ultrametric contour map produced by the gPb-owt-ucm method before applying the final postprocessing step; we denote this variant as $\mathbf{W}$-$\mathbf{Net}$+$\mathbf{ucm}$.   We can see that with this variant, our results outperform the other methods. 

Figure \ref{fig:no_ucm_result} illustrates results of running the proposed method $\mathbf{W}$-$\mathbf{Net}$ on images from the BSDS500. The first row shows the original inputs; the second row shows that results of initial boundaries detection produced by the $U_{Enc}$ encoder followed by a fully connected CRF. The third and the fourth rows show the ultrametric contour maps produced by the contours2ucm stage at the ODS and OIS respectively. Figure \ref{fig:add_ucm_result} illustrates results of running the $\mathbf{W}$-$\mathbf{Net}$+$\mathbf{ucm}$ on images from the BSDS500.



\section{Conclusion}

In this paper we introduced a deep learning-based approach for fully unsupervised image segmentation.  Our proposed algorithm is based on concatenating together two fully convolutional networks into an encoder-decoder framework, where each of the FCNs are variants of the U-Net architecture.  Training is performed by iteratively minimizing the reconstruction error of the decoder along with a soft normalized cut of the encoder layer.  As the resulting segmentations are typically coarse and over-segmented, we apply CRF smoothing and hierarchical merging to produce the final outputted segments.  On the Berkeley Segmentation Data Set, we outperform a number of existing classical and recent techniques, achieving performance near human level by some metrics.

We believe our method will be useful in cases where it is difficult to obtain labeled pixelwise supervision, for instance in domains such as biomedical image analysis where new data sets may require significant re-labeling for semantic segmentation methods to work well.  Further, our approach may be further refined in the future by utilizing different loss functions or postprocessing steps.  Ideally, we would like to design an architecture where additional postprocessing is not necessary.  Finally, designing variants of our architecture when a small amount of supervision is available would also be useful in many domains.

\section{Appendix}

\begin{figure*}
\begin{center}
\includegraphics[width=\textwidth]{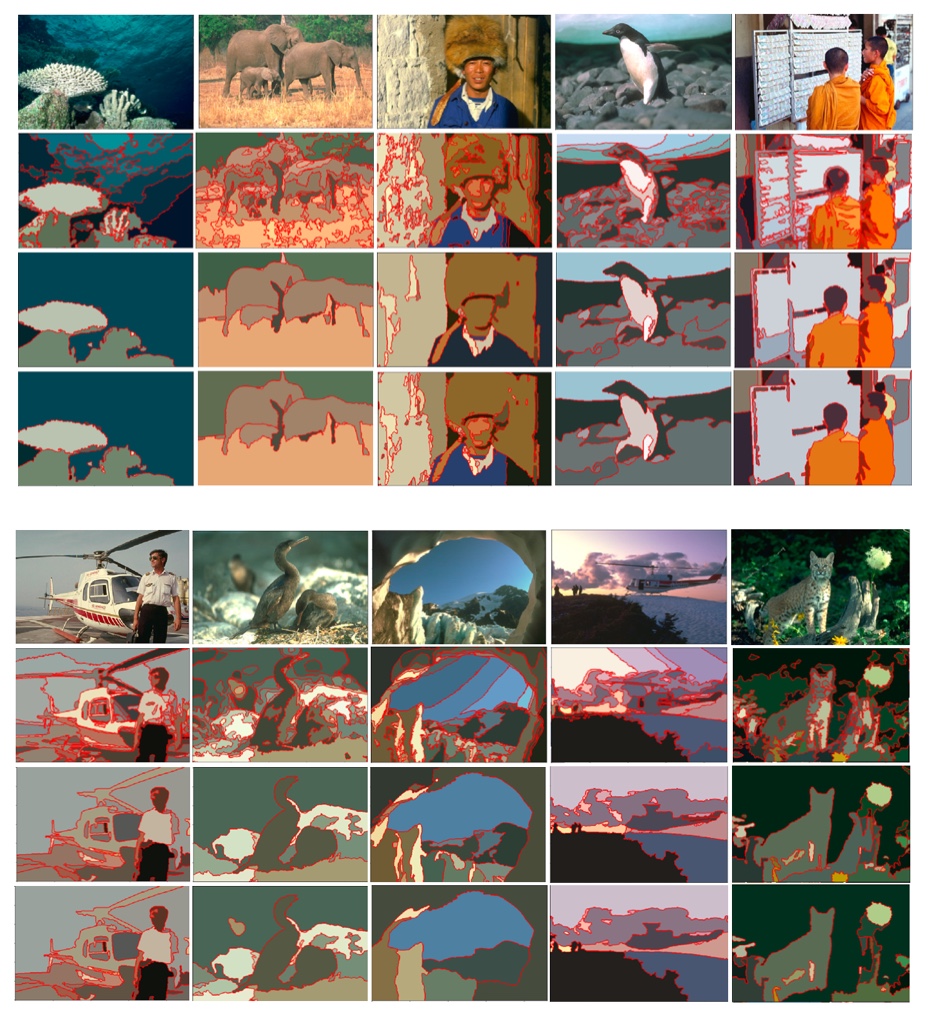}
\end{center}
   \caption{\textbf{Results of hierarchical segmentation using the output of WNet with CRF smoothing as initial boundaries, on the BSDS500.} From top to bottom: Original image, the initial over-segmented partitions showed in red lines obtained by the fully connected CRF, segmentations obtained by thresholding at the optimal dataset scale (ODS) and optimal image scale (OIS).}
\label{fig:no_horiz}
\end{figure*}

\begin{figure*}
\begin{center}
\includegraphics[width=\textwidth]{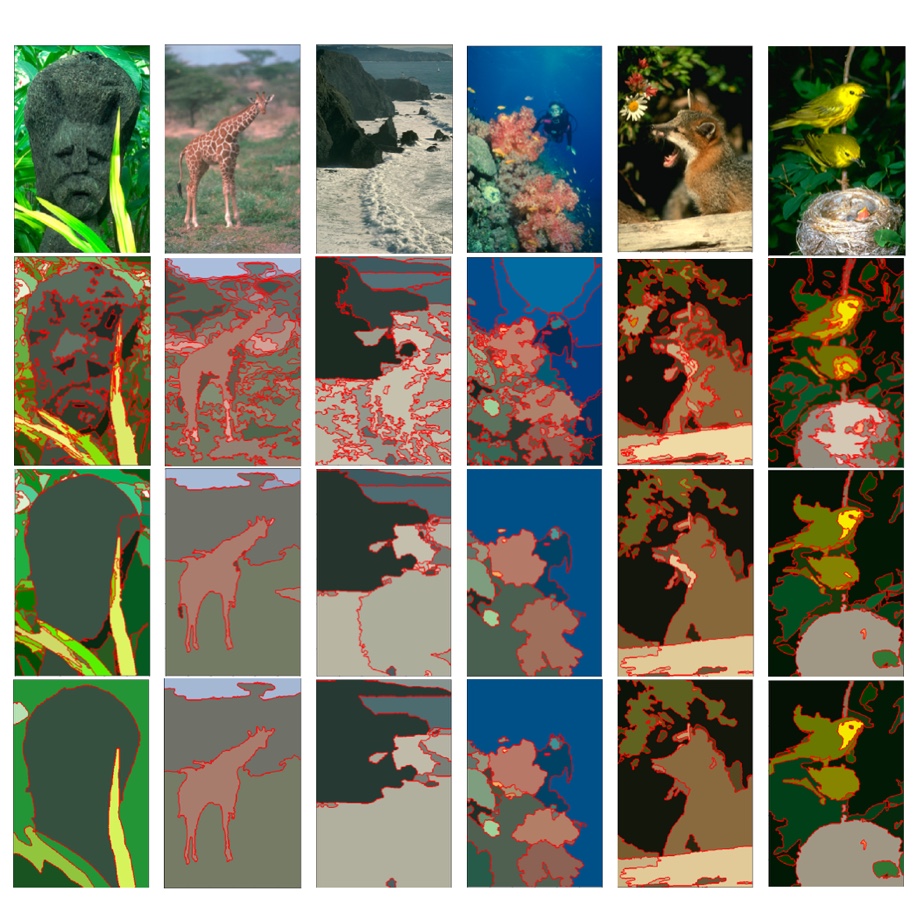}
\end{center}
   \caption{\textbf{Results of hierarchical segmentation using the output of WNet with CRF smoothing as initial boundaries, on the BSDS500.} From top to bottom: Original image, the initial over-segmented partitions showed in red lines obtained by the fully connected CRF, segmentations obtained by thresholding at the optimal dataset scale (ODS) and optimal image scale (OIS).}
\label{fig:no_vert}
\end{figure*}

\begin{figure*}
\begin{center}
\includegraphics[width=\textwidth]{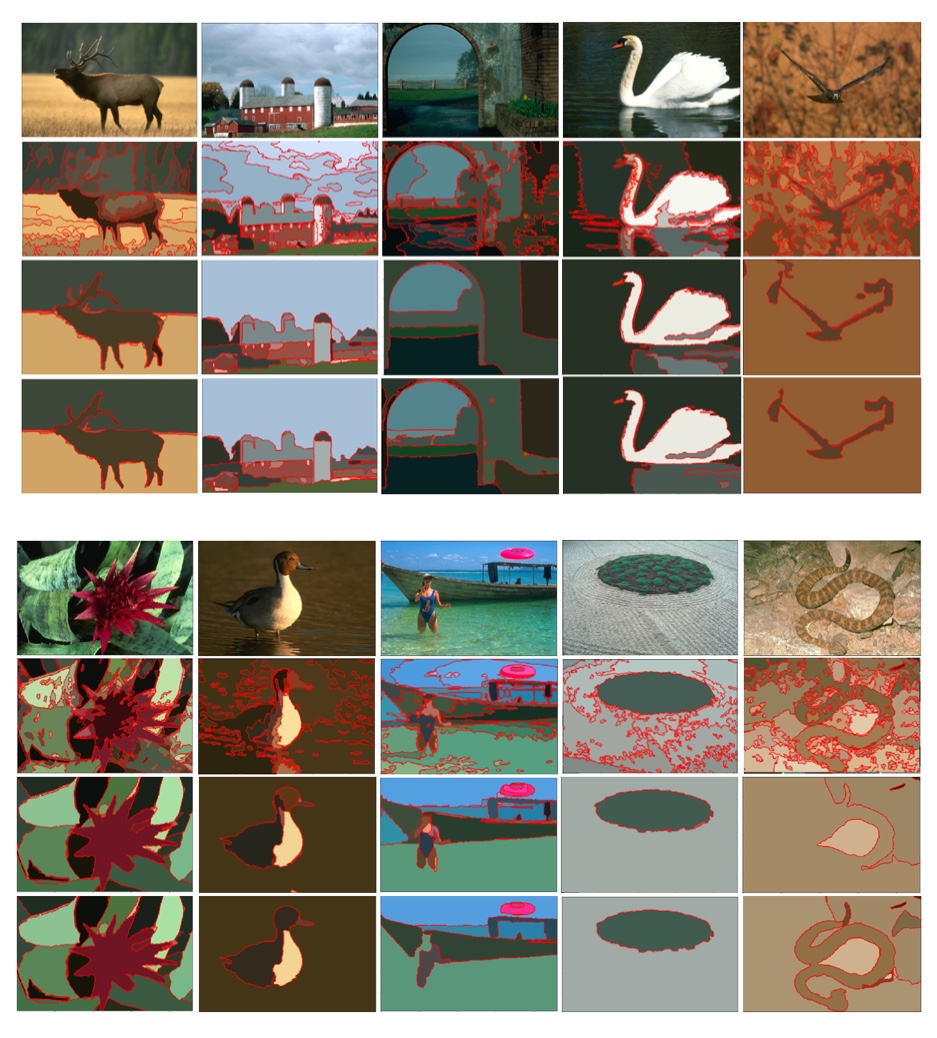}
\end{center}
   \caption{\textbf{Results of hierarchical segmentation using the combination of the output of WNet with CRF smoothing and UCM as the initial boundaries, on the BSDS500.} From top to bottom: Original image, the initial over-segmented partitions showed in red lines, segmentations obtained by thresholding at the optimal dataset scale (ODS) and optimal image scale (OIS).}
\label{fig:add_horiz}
\end{figure*}

\begin{figure*}
\begin{center}
\includegraphics[width=\textwidth]{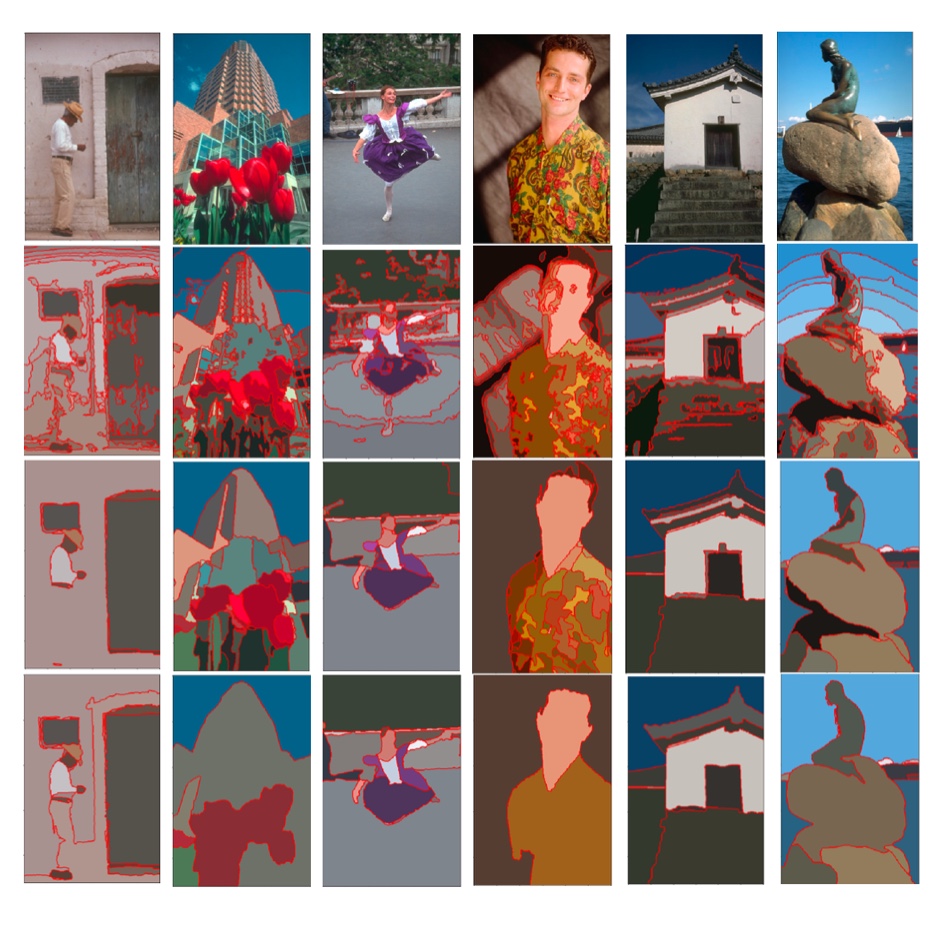}
\end{center}
   \caption{\textbf{Results of hierarchical segmentation using the combination of the output of WNet with CRF smoothing and UCM as the initial boundaries, on the BSDS500.} From top to bottom: Original image, the initial over-segmented partitions showed in red lines, segmentations obtained by thresholding at the optimal dataset scale (ODS) and optimal image scale (OIS).}
\label{fig:add_vert}
\end{figure*}

We show additional results of running the proposed method $\mathbf{W}$-$\mathbf{Net}$ on images from the BSDS500 in  Figure [\ref{fig:no_horiz}] and Figure [\ref{fig:no_vert}]. 
Further, Figure [\ref{fig:add_horiz}] and Figure [\ref{fig:add_vert}] illustrates more results of running the $\mathbf{W}$-$\mathbf{Net}$+$\mathbf{ucm}$ on images from the BSDS500.

{\small
\bibliographystyle{ieee}
\bibliography{egbib}
}

\end{document}